\documentclass{article}
\usepackage{graphicx} 
\usepackage{float}
\usepackage{amsmath}
\usepackage{tabularx}

\usepackage[colorlinks=true,linkcolor=blue,citecolor=blue,urlcolor=blue]{hyperref}

\usepackage[a4paper, left=1in, right=1in, top=1in, bottom=1in]{geometry}
\usepackage{amsmath, amssymb, amsfonts}
\usepackage{graphicx}
\usepackage{hyperref}

\title{\LARGE \textbf{Clothing agnostic Pre-inpainting Virtual Try-ON}}

\author{
  \textbf{Sehyun Kim}\textsuperscript{1}\thanks{Equal contribution} \quad
  \textbf{Hye Jun Lee}\textsuperscript{1}\footnotemark[1] \quad
  \textbf{Jiwoo Lee}\textsuperscript{1}\footnotemark[1] \quad
  \textbf{Taemin Lee}\textsuperscript{2}\thanks{Corresponding author} \\[6pt]
  \normalsize\textsuperscript{1}Department of Artificial Intelligence and Software, Kangwon National University, Korea \\
  \small\href{mailto:about7086@kangwon.ac.kr}{about7086@kangwon.ac.kr}, 
  \small\href{mailto:lovo6027@kangwon.ac.kr}{lovo6027@kangwon.ac.kr}, 
  \small\href{mailto:jiwo0723@kangwon.ac.kr}{jiwo0723@kangwon.ac.kr} \\[3pt]
  \normalsize\textsuperscript{2}Department of Electronic and AI System Engineering, Kangwon National University, Korea \\
  \small\href{mailto:kevinlee@kangwon.ac.kr}{kevinlee@kangwon.ac.kr}
}

\date{September 2025}
\begin{document}
\maketitle

\begin{abstract}
With the development of deep learning technology, virtual try-on technology has devel-oped important application value in the fields of e-commerce, fashion, and entertainment. The recently proposed Leffa technology has addressed the texture distortion problem of diffusion-based models, but there are limitations in that the bottom detection inaccuracy and the existing clothing silhouette persist in the synthesis results. To solve this problem, this study proposes CaP-VTON (\textbf{C}lothing \textbf{A}gnostic \textbf{P}re-Inpainting \textbf{V}irtual \textbf{T}ry-\textbf{On}). CaP-VTON integrates DressCode-based multi-category masking and Stable Diffu-sion-based skin inflation preprocessing; in particular, a generated skin module was in-troduced to solve skin restoration problems that occur when long-sleeved images are con-verted to short-sleeved or sleeveless ones, introducing a preprocessing structure that im-proves the naturalness and consistency of full-body clothing synthesis, and allowing the implementation of high-quality restoration considering human posture and color. As a result, CaP-VTON achieved 92.5\%, which is 15.4\% better than Leffa, in short-sleeved syn-thesis accuracy, and consistently reproduced the style and shape of the reference clothing in visual evaluation. These structures maintain model-agnostic properties and are appli-cable to various diffusion-based virtual inspection systems; they can also contribute to applications that require high-precision virtual wearing, such as e-commerce, custom styling, and avatar creation.
\end{abstract}


\section{Introduction}
With the rapid development of computer vision and deep learning technology, Con-trollable person image generation models have come to have important application value in the fields of e-commerce, fashion industry, and entertainment. In particular, virtual try-on technology is contributing to improved purchase satisfaction and reduced return rates by allowing consumers to check the wearing effect in advance without actually wearing clothes in an online shopping environment. Against this background, the Learn-ing Flow Fields in Attention(Leffa) method proposed by Zhou et al.\cite{Zhou2025} is drawing attention as an innovative approach to solving the detailed texture distortion problem that existing diffusion-based models were experiencing.

Existing controllable person image generation methods have achieved a high level of overall image quality, but have still shown limitations in accurately preserving fine tex-ture details from reference images. These issues have been particularly noticeable in clothing containing complex patterns, logos, and texts, and have acted as important bar-riers to implementing practical virtual inspection services. Leffa identified the root cause of this problem as the attention mechanism's inability to focus on the correct area of the reference image, and addressed this by introducing a normalization loss that induces the learning of the flow field within the attention layer to focus explicitly on the correct refer-ence area.

\begin{figure}[htbp]
    \centering
    \begin{minipage}{0.49\linewidth}
        \centering
        \includegraphics[width=\linewidth]{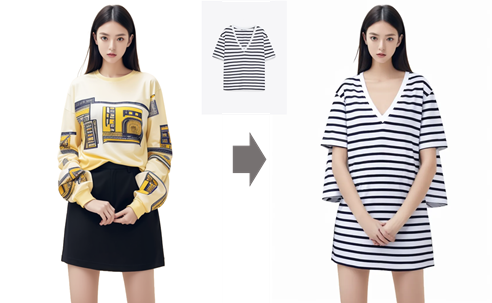}
    \end{minipage}
    \begin{minipage}{0.49\linewidth}
        \centering
        \includegraphics[width=\linewidth]{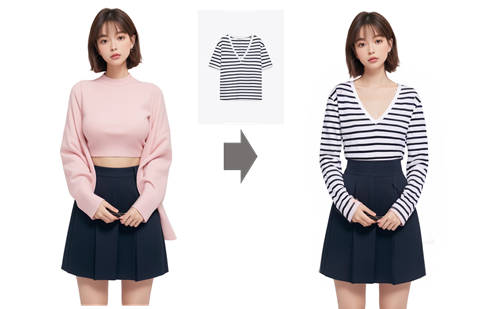}
    \end{minipage}
    \caption{Limitations of previous Leffa model: (a) Problem of detecting bottom, (b) Synthesis limitations dependent on source input images}
    \label{fig:placeholder}
\end{figure}

However, there were synthesis results that could not be covered by the existing Leffa model, and we supplemented them by adding several things from the pipelines of the existing study. The points that can be supplemented by the existing Leffa model are shown in Figure 1. The first can be confirmed in Figure 1(a). Since the existing Leffa model is mainly optimized for virtual inspection centered on upper-body clothes, detection and masking of the lower-body areas are relatively vulnerable. This can put great restrictions on this part in a virtual inspection scenario centered on whole-body clothing or bottoms. In particular, when the lower-body could not be detected in the input image, there was a limitation in that the whole body was synthesized even if the upper-body clothes were synthesized. Therefore, a separate specialized preprocessing strategy that elaborately reflects this is needed. The second was that the existing clothing characteristics included in the input source image had more than a certain level of influence on the final output result. Figure 1(b) was outputted as a Leffa-Corrective Masking version that improved the first limitation to confirm independently of the first problem. When trying to synthesize short-sleeved clothing with the image of a person wearing a long-sleeved top as shown in Figure 1(b), it was observed that the existing long-sleeved shape was maintained regardless of the input garment. This seemed to be because the skin area to be exposed, such as the forearm, shoulder, and neck, was not sufficiently restored in the process of removing clothing, and the existing clothing type acted as an interference factor in the subsequent clothing synthesis process. This phenomenon could lead to unnatural results in the synthesis between clothing with large differences in shape such as sleeve length and neckline.

By integrating the Dress Code-based improved masking mechanism and skin inpainting pipeline while maintaining the advantages of the existing Leffa pipeline, this study aims to achieve complete synthesis results in all clothing categories and bottoms. This, combined with Leffa's excellent attention-based flow field learning ability, was able to present an integrated solution that satisfies the three goals at the same time: detailed texture preservation, full-body clothing synthesis, and input image independent clothing replacement. In particular, it was confirmed that the interference phenomenon caused by existing clothing characteristics was significantly improved along with the morphological inconsistency, the unnaturalness of the boundary, and the incongruity with the body ratio shown in the bottom synthesis. In addition, this study is of high practical value in that it provides a universal improvement method that can be applied to various diffusion-based models through improved accuracy of the masking process and improved skin restoration quality while maintaining the model-agnostic characteristics of the existing Leffa model. This approach is differentiated from existing studies that have presented solutions dependent on specific models or datasets, and is expected to be widely used in various applications in the virtual try-on field in the future.

\clearpage

\section{Related Work}
\subsection{Evaluation of Generative Models}
The generative model is difficult to evaluate compared to the classification model. This is because there is no single image to compare with the results obtained through the generative model. It is necessary to create an index that can evaluate following the distri-bution of actual domain images, and various evaluation indicators have been proposed. The indicators used in the evaluation are as follows.

FID(Frechet Inception Distance) is a key indicator of the quality of images created by generative models\cite{Heusel2017}. FID measures the statistical distance between the two distributions by comparing the distribution of real and generated images, extracts feature vectors in 2,048 dimensions using a pre-trained Inception v3 model, and measures the distance by calculating the mean ($ \mu $) and covariance (C) of the real and generated images. This means that the lower the FID value, the more similar the generated image is to the real image.

\[
d^2 = \|\mu_1 - \mu_2\|^2 + \operatorname{Tr}\left( C_1 + C_2 - 2\left( C_1 C_2 \right)^{1/2} \right)
\]

SSIM(Structural Similarity Index Measure) is an index that quantifies structural sim-ilarity between two images\cite{Wang2004}. SSIM outputs a value between 0 and 1 by comparing the luminance, contrast, and structural elements of two images, and a closer to 1, the more structurally similar the two images are. In particular, SSIM is more advantageous in eval-uating the degree of retention of patterns and local structures than errors in pixel units. Generally, it is used to evaluate the visual quality of high-resolution images.

\[
SSIM(x, y) = \frac{(2\mu_x\mu_y + C_1)(2\sigma_{xy} + C_2)}{(\mu_x^2 + \mu_y^2 + C_1)(\sigma_x^2 + \sigma_y^2 + C_2)}
\]

LPIPS(Learned Perceptual Image Patch Similarity) is an indicator that quantifies perceptual differences between two images using deep learning-based image recognition models\cite{Zhang2018}. LPIPS measures the similarity by extracting feature vectors from different lay-ers of pre-trained neural networks (e.g., AlexNet, VGG, SqueezeNet, etc.). The lower the LPIPS value, the more visually similar the two images are, and it has a complementary relationship with FID and SSIM in that it better reflects high-dimensional semantic simi-larity than simple pixel differences.

\[
\mathrm{LPIPS}(x, y) =
\sum_{l}
\frac{1}{H_l W_l}
\sum_{h,w}
\left\| w_l \odot \left( \hat{f}_l^x(h, w) - \hat{f}_l^y(h, w) \right) \right\|_2^2
\]

The various indicators described in this chapter are used to compare the results of the model we proposed in 4.2.2 with existing models. In addition to this, in this study, the form-consistency of the virtual inspection model is regarded as a key criterion directly re-lated to the actual service quality, and a statistical-based evaluation index was introduced to measure this objectively. The normal output rate is a representative statistical-based quantitative indicator, which refers to the percentage (\%) of cases where the normal short-sleeved silhouette is accurately synthesized, especially when the image of wearing long-sleeve clothing is input and the reference clothing is short-sleeve. The main purpose of this indicator is to directly quantify the degree of reflection of the reference clothing sil-houette and the interference phenomenon of input clothing characteristics, which are dif-ficult to explain with simple image quality and similarity indicators (FID, SSIM, LPIPS, etc.). Existing models often did not match the actual reference clothing type (e.g., short-sleeve, sleeveless, etc.) and left the input clothing silhouette (especially long-sleeve), but the proposed technique was fundamentally improved through pipeline strategies such as generate skin and improvement masking. The normal output rate calculation is performed by evaluating the output image when the reference clothing is short-sleeved for long-sleeved input images in the dataset, classifying whether the short-sleeved form is actually implemented (normal/abnormal) as in the example, and then calculating the normal ratio.

\clearpage
\subsection{Virtual Try-on Models}
Virtual Try-on models can be divided into classical models, diffusion-based models, and attention-based models. Early Virtual Try-On approaches mainly used GAN-based warping techniques and segmentation. CP-VTON\cite{WangECCV2018} handled clothing deformation using thin plate spline (TPS) warping and segmentation, but there was a limit to realistic image generation as the FID value was very high at 47.36. Since then, VITON-HD\cite{Choi2021} has achieved FID 11.74 at 1024×768 resolution by introducing alignment-Aware segment (ALIAS) nor-malization and warping unit, but there was still a limit to accurate clothing alignment and natural deformation in complex poses. ACGPN\cite{Yang2020} introduced an adaptive content generating and preserving network to record an FID of 26.45, but its limited performance at 256×192 resolution made it difficult to expand high resolution, and HR-VITON\cite{LeeECCV2022} im-proved to FID of 10.91 afterwards, but the sleeve-squeezing and waist-squeezing artifact problems persisted. These classical models show an average FID of 24.06, with funda-mental limitations in preserving complex clothing textures and generating realistic wear-ing effects.

With the advent of diffusion models from 2023, great progress has been made in the field of Virtual Try-On. Diffusion models enabled more stable learning and high-quality image generation compared to GAN(Generative Adversarial Network)\cite{Goodfellow2014}, and achieved a lot of performance improvement compared to the past with an average FID of 7.18. Sta-bleVITON\cite{Kim2024} achieved FID 6.52 by applying zero cross-attention to LDM (Latent Diffu-sion Model), but there was a segment detail loss problem due to attrition misalignment. IDM-VTON\cite{Li2024} improved to FID 6.29 by utilizing segment-agnostic person representation and mask, but there was a limit to customization deterrence and continuous consistency in complex clothing transformation. LaDI-VTON\cite{Morelli2023} recorded a relatively high fid value with an FID of 8.85 by combining Latent Diffusion, Textual-Inversion, and CLIP. OOTDif-fusion \cite{Xu2024} implemented multiple outfit processing through parallel outfitting UNet and outfitting fusion, and recorded an average fid score of about 10 with 11.03 on top, 9.72 on bottom, and 10.65 on dress for each part.

Later, the introduction of transformers and attention mechanisms enabled more so-phisticated feature matching and spatial relationship modeling even in Virtual Try-On. The multi-head self-attention proposed in Vaswani, A. et al.\cite{Vaswani2017} became the basis of the transformer model, and models that use it in Virtual Try-On also began to emerge. Cat-VTON\cite{Chong2025} selectively fine-tuned only the self-attention module within the diffusion U-Net and conducted lightweight learning with 49.57M parameters, removing text encoders and cross-attention, and maximizing efficiency by simply concatenating and inputting human and costume images in the spatial dimension. However, detailed semantic alignment control was limited. 

TryOnDiffusion\cite{Zhu2023} achieved precise garment-person matching by leveraging garment-aware person representations and cross-attention based on latent diffusion. It recorded an FID of 23.3 on the VITON-HD dataset, demonstrating improved performance over existing diffusion models in terms of detailed clothing texture reproduction and pose preservation. HumanDiffusion\cite{Zhang2023-1} proposed a diffusion pipeline that separates human image representations into a distinct latent space and combines cross-attention with mask inpainting. It delivered natural results for full-body synthesis and multi-clothing category replacement, achieving FID scores ranging from 30.4 to 31.2 on the DeepFashion-Multimodal dataset.

Leffa\cite{Zhou2025} improved the self-attention and cross-attention map by ap-plying flow field regularization to the attachment of the diffusion model. This enabled more sophisticated attention control than the existing mask inpainting method, zero cross-attention method, and prompt-aware mask method, but there were still problems with the complexity and fine-detail distortion of flow field regularization.

Existing Leffa applied flow field regularization unique to the attention mechanism to enable more sophisticated attention control than existing methods, but there are still limi-tations in complex clothing deformation, detailed texture preservation, and accurate area segmentation in the pre-processing stage. This was a situation as shown in Figure 1, and several steps were added to the model's pipeline to overcome this limitation.

In this study, we propose an inpainted pre-processing input-based attention flow system to improve the limitations found by existing Leffa model. We propose the CaP-VTON(\textbf{C}lothing \textbf{A}gnostic \textbf{P}re-Inpainting \textbf{V}irtual \textbf{T}ry-\textbf{On}) model by adding a part that can compensate for the shortcomings to the existing VITON-HD model of the Leffa model. Through this model, the stability of the diffusion model, the attention mechanism of the transformer, and the advantages of the classical warping technique are selectively integrated to solve the problems of the limitations of the attention control, the inaccuracy of the pre-processing step, and the lack of consistency for various clothing types of existing studies.
\clearpage
\begin{figure}
    \centering
    \includegraphics[width=1\linewidth]{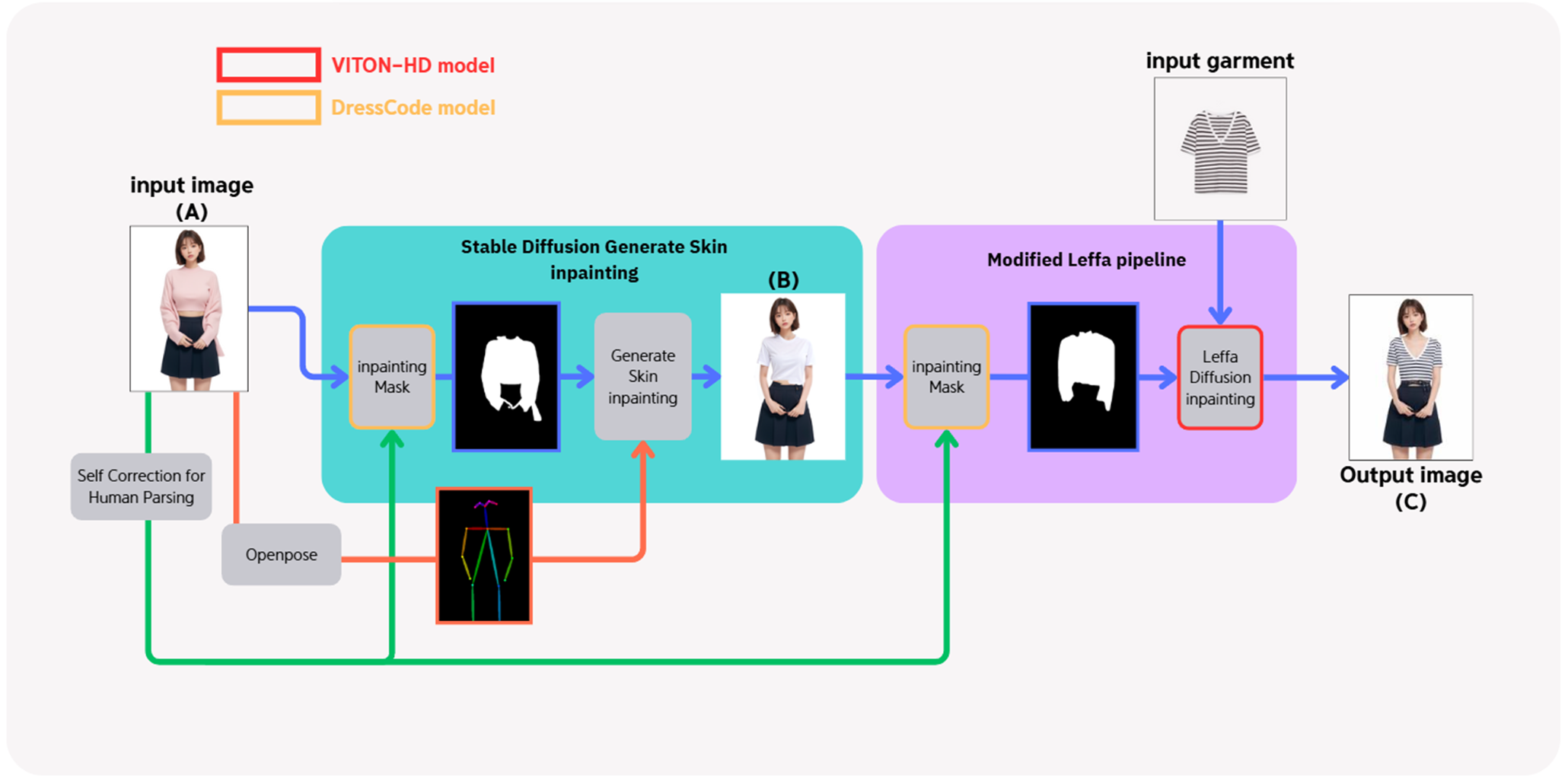}
    \caption{Proposed method: CaP-VTON(Clothing agnostic Pre-inpainting Virtual Try-ON)}
    \label{fig:placeholder}
\end{figure}

\section{Proposed Method}
\subsection{System Flow: Concept of CaP-VTON}
This study proposes the CaP-VTON(\textbf{C}lothing \textbf{A}gnostic \textbf{P}re-Inpainting \textbf{V}irtual \textbf{T}ry-\textbf{On}) model, a new model that integrates two key improvement factors to complement the structural limitations of the existing Leffa pipeline and enable a natural and coherent transition between various clothing categories. The proposed improvement factors are the Dress Code-based multi-category masking stage, which covers full-body clothing such as top, bottom, and dress, and the Stable Diffusion-based skin inpainting pre-processing stage for removing existing clothing silhouettes and restoring exposed skin.

The overall flow is shown in Figure 2. This study can be largely divided into two stages. One is the Modified Leffa Pipeline for the limitation that detection and masking of the lower-body cloths area is relatively weak because the existing VITON-HD Leffa model is mainly optimized for virtual inspection centered on upper-body cloths, and the other is the Stable Diffusion Generation Skin Inpainting step for the limitation that the existing clothing characteristics included in the input source image affect the final output result more than a certain level. In the first step, when the user inputs a source image, the se-mantic part of the human body is first subdivided using the SCHP(Self-Correction for Human Parsing) model\cite{Li2021}, and at the same time, various clothing areas such as top, bot-tom, and dress are accurately masked through the Dress Code model. In addition, by ex-tracting the person's posture information using OpenPose\cite{Cao2017}, the human structure is not distorted during the subsequent inpainting process. The body mask and pose information obtained in this step are used in the inpainting process to remove the existing clothing ar-ea and naturally restore the skin and short-sleeved silhouette. In the first step, an pre-processed inpainting image(B in Figure 2) is generated from the input source image(A in Figure 2). In the second step, a mask of the pre-processed image is made through the Dress Code model by using the pre-processed inpainting image(B in Figure 2) output in a clean form without the silhouette of the existing long-sleeved clothing, and is transmitted to diffusion painting applying the attachment-based flow field regularization of the VI-TON-HD model along with the reference clothing information to generate a final virtual fitting image(C in Figure 2).

\begin{figure}[H]
    \centering
    \begin{tabular}{cc}
        \includegraphics[width=0.45\linewidth]{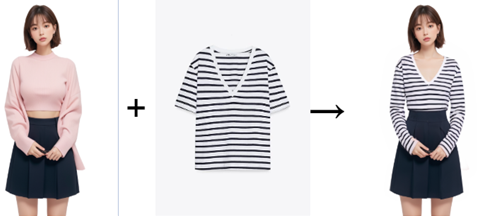} &
        \includegraphics[width=0.45\linewidth]{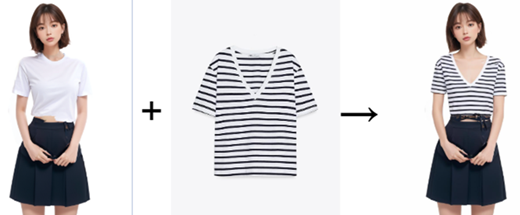} \\
        \small{(a)} & \small{(b)} \\
        \includegraphics[width=0.45\linewidth]{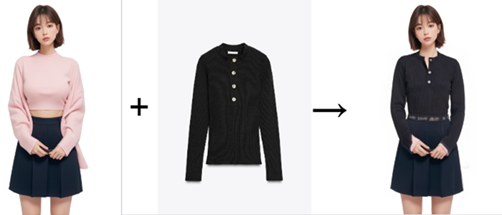} &
        \includegraphics[width=0.45\linewidth]{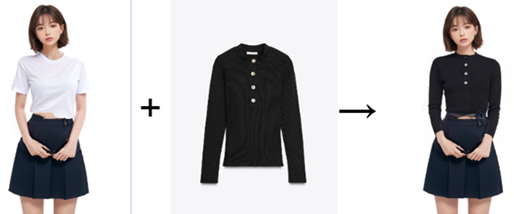} \\
        \small{(c)} & \small{(d)}
    \end{tabular}
    \caption{Changes in output results according to Leffa clothing type.}
    \label{fig:leffa-clothing-type}
\end{figure}

\subsection{Stable Diffusion Generate Skin Inpainting: Removing existing clothing silhouette}
When trying to synthesize short-sleeved clothing from the image of a person wearing a long-sleeved top using the existing model, a problem arises in that the existing long-sleeved shape is maintained regardless of the type of reference clothing, as shown in Figure 3(A) and (C), and the silhouette of the existing clothing remains, as shown in Figure 3(B). This is because the silhouette of the existing clothing interferes with the synthesis of new clothing because the areas of exposed skin, such as the forearms, shoulders, and neck, are not properly restored during the clothing removal process.

To solve this problem, the generate skin method pipeline inserts short-sleeved clothes and realistic skin into the area where existing clothes have been removed, which does not affect the output. This is intended to accurately reflect the unique shape and style of the reference clothes regardless of the existing clothing characteristics of the input image.

The number of data points used in the garment silhouette removal experiment in the Generate Skin interposing module was set to 599 to simultaneously secure the reproduci-bility and the efficiency of the experiment in a limited GPU resource environment (Colab-based). “Wearing Held Tight Short Sleeve Shirt, high quality skin, realistic, high quality" was used for the prompt, and "Blurry, low quality, artifacts, deformed, ugly, tex-ture, watermark, text, bad anatomy, extra limbs, face, hands, fingers" were used to pre-clude the creation of elements corresponding to the prompt. The inference step was set to 20 steps, and no separate fine-tuning was performed.

\begin{figure}[htbp]
    \centering
    \includegraphics[width=0.7\linewidth]{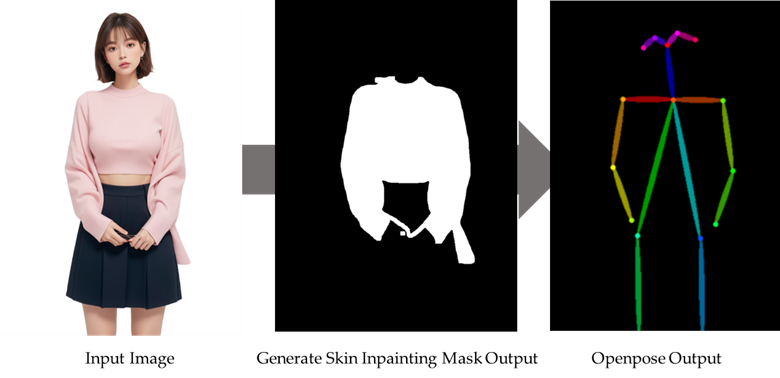}
    \caption{Preprocessed Images for Stable Diffusion Generate Skin inpainting}
    \label{fig:placeholder}
\end{figure}
\subsubsection{Generating input information to build synthetic model}
To perform inpainting through the Generate Skin function, an OpenPose image con-taining the inpainting mask image and the pose information of the input image is re-quired. The Generate Skin Inpainting Mask Output of Figure 4 was generated using the masking pipeline provided by Leffa's DressCode model. In addition, the OpenPose Output of Figure 5 has been used as conditioning information for the ControlNet in the process of Stable Diffusion-based inpainting. As shown in Figure 5, when the ControlNet is not used, the pose identity of the original person is not accurately transmitted to the network, re-sulting in the problem of creating a structurally distorted image, featuring, for example, a misalignment of the body position such as the arm or shoulder, or mixing with the back-ground. As a result, we determined that the problem of preserving the pose and securing the accuracy of the skin area in the process of inpainting directly affects the realism of the final result, and to solve this problem, information on the Generate Skin Inpainting Mask and OpenPose-based ControlNet [20] was used as the input information for the Generate Skin function. In addition, as can be seen in Figure 5, there were cases wherein the color of the inpainted skin did not properly reflect the skin tone of the original image, creating darker or brighter skin than was actually present. To compensate for this, the Skin Tone Extraction technique was introduced to create a skin texture that naturally harmonizes with that in the original image. This technique is implemented by detecting the skin area in the YCrCb color space and calculating the average skin tone value in the HSV color space. The skin area created through this approach is naturally blended with the skin tone of the original image, such that the overall visual consistency of the resulting image can be maintained.

\begin{figure}[htbp]
    \centering
    \includegraphics[width=0.8\linewidth]{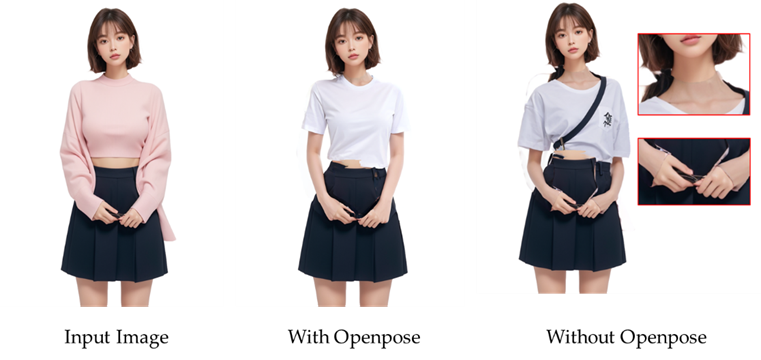}
    \caption{Comparison of results with/without Openpose ControlNet}
    \label{fig:placeholder}
\end{figure}

\subsubsection{MajicMix Realistic model based high quality skin synthesis}
The core of the Generate Skin pipeline is inpainting the skin area using the MajicMix Realistic model\cite{Merjic2023}. This model is based on Stable Diffusion 1.5, and is a high-quality checkpoint model trained by fusing various model weights. In particular, it is optimized for generating Asian skin tones, so it has a strength in deriving natural results from Asian images. In addition, this model introduces Noise Offset Technique, which enables more sophisticated light source expression and smooth shadowing during the diffusion process. It exhibits excellent performance in restoring texture, shade, and color of complex skin ar-eas such as arms, shoulders, and neck exposed after clothing removal. Figure 6 shows the results output through the generate skin pipeline. 

\begin{figure}[H]
    \centering
    \includegraphics[width=0.8\linewidth]{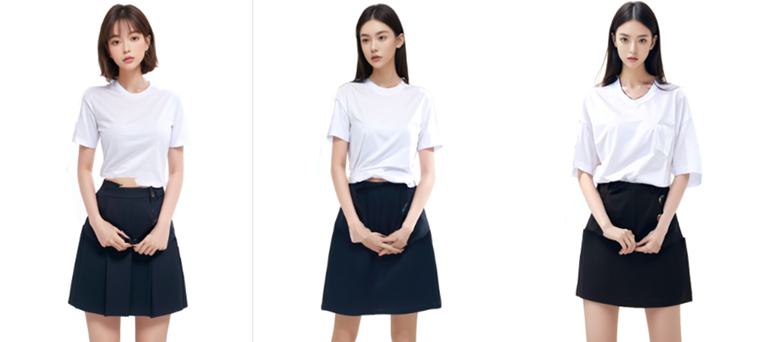}
    \caption{Examples of results of generate skin pipeline}
    \label{fig:placeholder}
\end{figure}

\begin{figure}[htbp]
    \centering
    \includegraphics[width=0.8\linewidth]{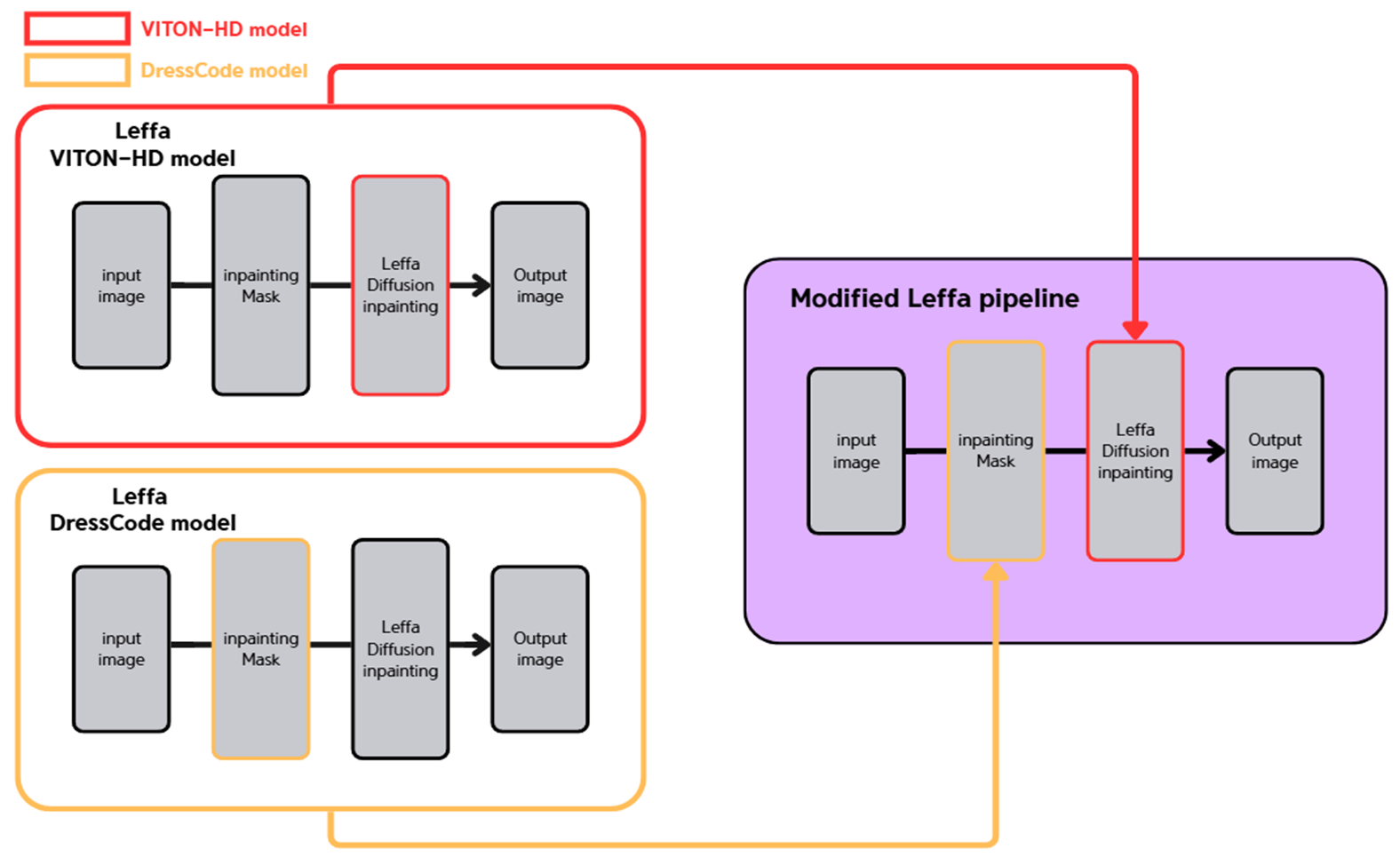}
    \caption{Modified Ensemble model of VITON-HD and DressCode Model}
    \label{fig:placeholder}
\end{figure}
\subsection{Modified Leffa pipeline: Multi-category enabled based on DressCode}
There are two models in the Leffa model: one trained with VITON-HD data and one trained with DressCode data. The VITON-HD Leffa model was optimized for phase-oriented synthesis. For this reason, there were limitations in the detection of the lower area and the whole body synthesis, and the image with a high proportion of the lower body showed unstable synthesis results. To solve this problem, as shown in Figure 7, the inpainting Mask used the DressCode model, and the one of the VITON-HD model with high output image quality was ensembled in the Leffa Diffusion inpainting output. A masking strategy was constructed using the multi-category structure of the DressCode dataset.

\begin{figure}[H]
    \centering
    \begin{minipage}{0.49\linewidth}
        \centering
        \includegraphics[width=0.85\linewidth]{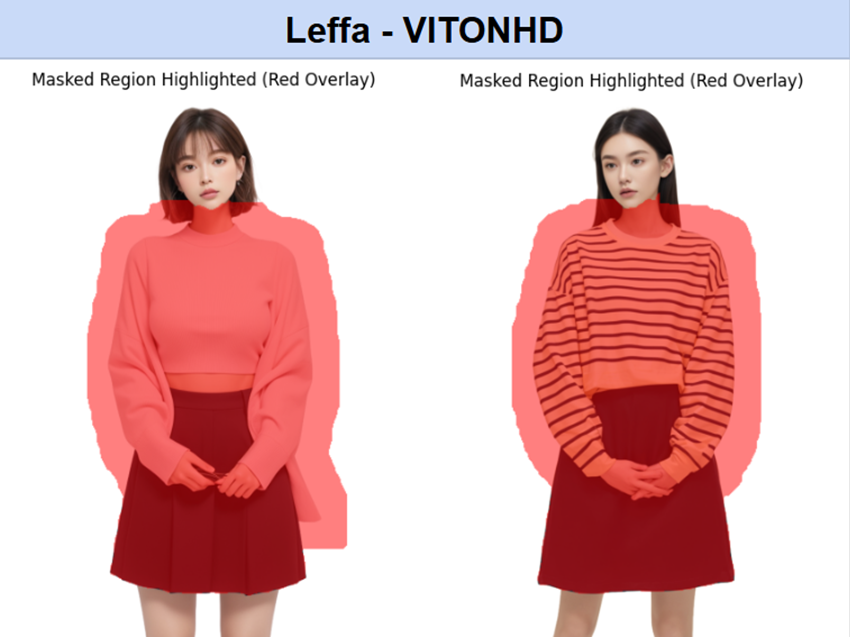}
    \end{minipage}
    \begin{minipage}{0.49\linewidth}
        \centering
        \includegraphics[width=0.85\linewidth]{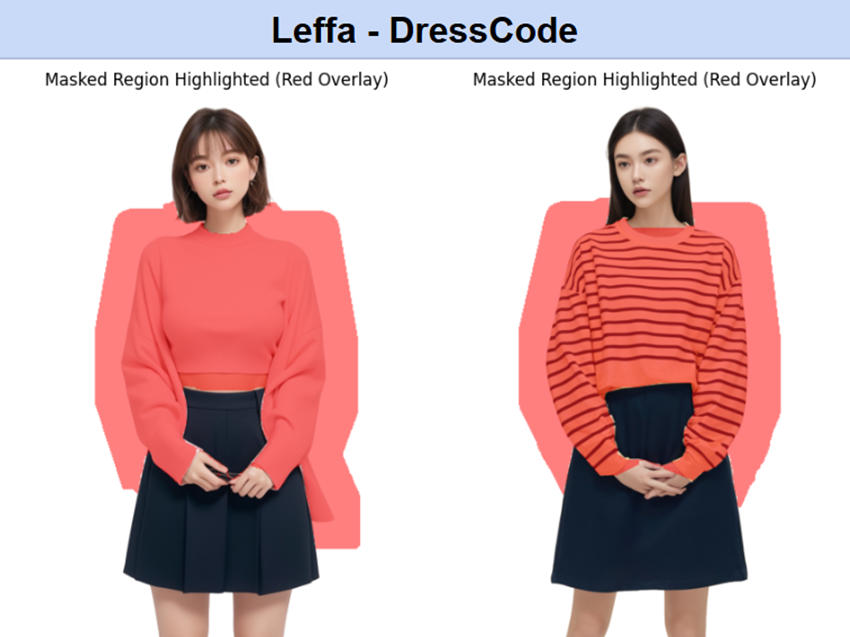}
    \end{minipage}
    \caption{Comparison of two masking methods: Red colored highlighted regions for masking}
    \label{fig:placeholder}
\end{figure}

\subsubsection{Characteristic of VITON-HD and DressCode Model}
Figure 8 shows the comparison of the masking quality of the two models. The VI-TON-HD model is trained on the VITON-HD dataset specializing in upper body clothing images and is mainly optimized for image synthesis. Accordingly, the image centered on the upper body shows excellent visual quality, but the information on the lower body is not included, so there is a limit to the detection and masking of the lower body area. On the other hand, the Dress Code model was learned based on a dataset including various categories of clothing such as tops, bottoms, and dresses, which allows masking for full-body clothing. In fact, in this study, by using the Dress Code model when generating an inpainting mask, stable masking results were secured in multi-category situations.

To compensate for these structural differences, this study designed a pipeline in a way that combines the advantages of the two models. In the in-painting masking process, multi-category detection was performed using the Dress Code model, and better image quality was secured by using the VITON-HD model in the final Diffusion synthesis step. To verify this numerically, the synthesis performance of the two models was compared as shown in Table 1. As a result, the FID score was 4.54 for the VITON-HD model, which was lower than the Dress Code model (5.05), showing an advantage in terms of the overall visual quality of the generated image. On the other hand, in the SSIM and LPIPS indicators, the Dress Code model is better, showing that it is superior in terms of structural and de-tailed texture expression of actual clothes.

\begin{table}[htbp]
    \renewcommand{\arraystretch}{1.5} 
    \centering
    \caption{Numerical Comparison between two models (paired : If the costume in the input person image and the costume image provided as a reference are the same, the result of the production can be compared directly to the correct image because the model actually produces the image in the same clothes. Therefore, we used quantitative image quality indicators such as SSIM and LPIPS in addition to the FID. unpaired : refers to the case where the costume in the input person image and the costume image given as reference are different. In this case, since there is no correct answer for the generated image, indicators such as SSIM or LPIPS cannot be used. Instead, we only used dis-tribution-based indicators such as FID, which can evaluate the overall quality of generation. In-dicators with down arrows have higher similarity with lower values, and indicators with up ar-rows have higher similarity with higher values.}
    \begin{tabularx}{\textwidth}{l|>{\centering\arraybackslash}X>{\centering\arraybackslash}X>{\centering\arraybackslash}X|>{\centering\arraybackslash}X}
        \hline
        \textbf{Model} & \textbf{FID $\downarrow$} & \textbf{SSIM $\uparrow$} & \textbf{LPIPS $\downarrow$} & \textbf{FID $\downarrow$} \\
        \hline
        VITON-HD    & 4.54 & 0.899 & 0.048 & 8.52 \\
        DressCode   & 5.05 & 0.949 & 0.021 & 10.73 \\
        \hline
    \end{tabularx}
    \label{tab:numerical_comparison}
\end{table}

\section{Experimental Results}
\subsection{Experimental Environment}
This study was conducted in the Google Colab environment, and Ubuntu 20.04 LTS was used as an operating system. The programming language consisted of Python 3.11.13 version, and GCC 11.4.0 was used as the compiler. As the main library, packages related to image processing and deep learning such as OpenCV and TorchGeometry were used, centering on PyTorch (1.6.0 or higher, support for GPU acceleration).

As a generative model for image generation, the ChilloutMix checkpoint based on Stable Diffusion 3.5 was used. This is a derivative weight (checkpoint) that improves hu-man texture and lighting expression while maintaining the structure of Stable Diffusion 3.5, and was driven through the Diffusers library. The main settings in the generation process consisted of cfg scale = 7, steps = 25, sampler = Euler a, clip skip = 2. Person imag-es generated through this setting were used for matching and synthesis experiments with clothing images within the CaP-VTON pipeline.

The hardware configuration consisted of NVIDIA A100 GPU (40GB memory), 83.5GB system RAM, and Intel Xeon-based CPU, and the experiment was conducted through the Colab interface based on Jupyter Notebook. When the Stable Diffusion model was config-ured with the above settings, it took an average of about 8 seconds to generate a single image.

To assess whether the model generalizes effectively to new data and maintains stable performance on real human images, two types of person data were prepared. Specifically, 100 AI-generated person images were produced using the ChilloutMix model, and 1,000 real person images were obtained from the VITON-HD Test dataset. These were combined to create a unified person dataset containing 1,100 images, paired with a cloth dataset of equal size (1,100 garment images).

All images were normalized to a base resolution of 1024×768 to ensure reproducibility and computational efficiency under the limited GPU environment provided by Colab. As a result, the final experimental dataset consisted of 1,100 person images and 1,100 cloth images.

\begin{figure}[H]
    \centering
    \includegraphics[width=1\textwidth]{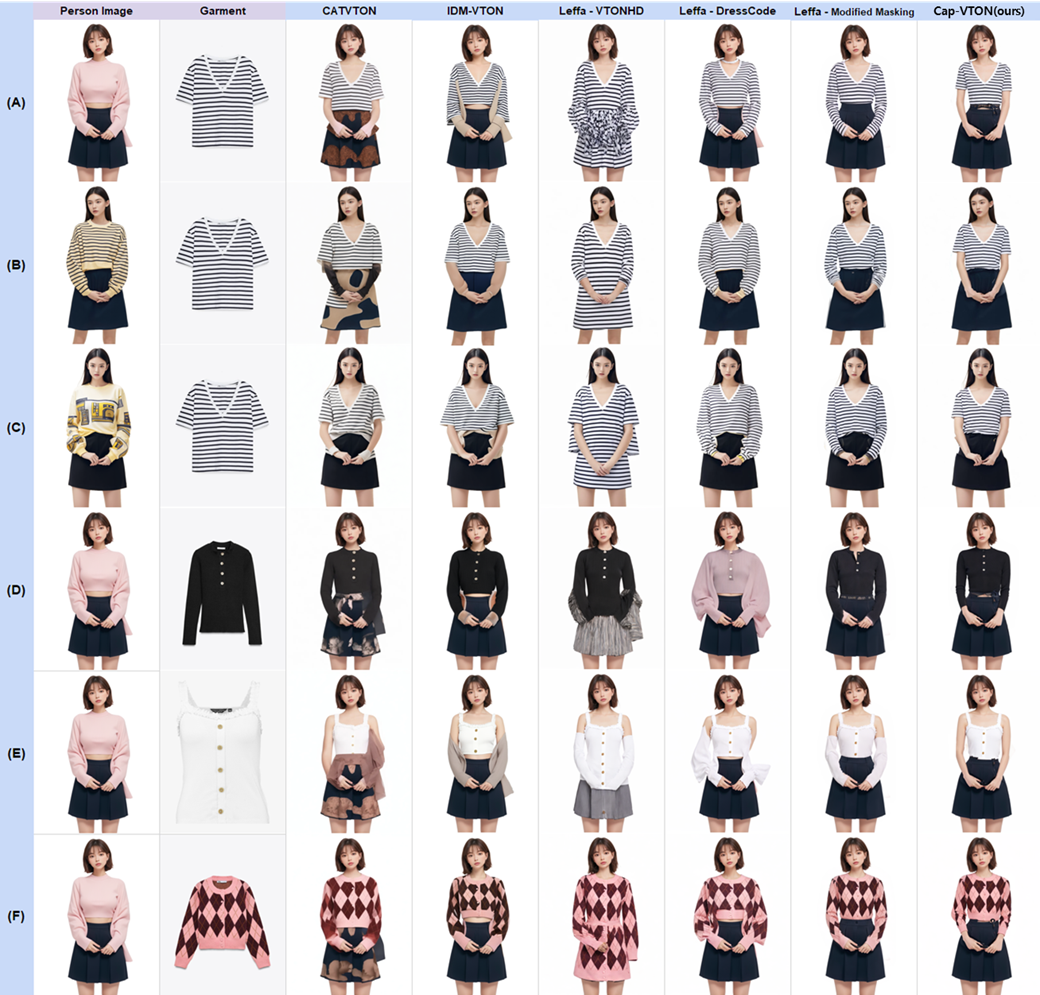}
    \caption{Result of our method (on last column): 1st column is input person image, and 2nd images are the clothes for change. 3rd--7th columns are synthesized results of other methods.}
    \label{fig:placeholder}
\end{figure}
\subsection{Results}
\subsubsection{Visual Comparison of the synthesis results}
Figure 9 is a visual comparison of the synthesis results generated by various virtual wearing models based on the same input image and the same reference clothing. This comparison was evaluated beyond simple visual quality, focusing on whether the silhou-ette of the reference clothing (especially the sleeve length and arm exposure area) was ac-curately reflected.

As a result of the comparison, the existing CatVTON, IDM-VTON, Leffa-VITONHD, and Leffa-DressCode models were commonly influenced by the existing long-sleeved sil-houette of the input image, showing limitations in not being able to accurately reproduce the design of reference clothing, such as abnormally long sleeves in the output image or insufficient exposure of arms and shoulders. In particular, it was observed that the Leffa-VITONHD model was synthesized up to the lower half, which should not be syn-thesized, and despite the fact that short-sleeved clothing was actually applied, it was ob-served that the shape of the existing long-sleeved sleeves remained on the output or the human body was unnaturally distorted.

On the other hand, the proposed CaP-VTON (Ours) model effectively removes exist-ing long-sleeved clothing and naturally and consistently restores skin areas such as arms, shoulders, and armpits to be exposed by utilizing the generate skin-based preprocessing technique. As a result, the short-sleeved silhouette of the reference clothing was clearly implemented in the output image, and the pattern and shape were neatly reflected with-out colliding with the existing clothing. This difference does not stop at just the visual naturalness, but acts as an important indicator for evaluating how accurately the style of the reference clothing can be reflected. While existing models are strongly influenced by input clothing, CaP-VTON shows that it is possible to reproduce a form independent of input clothing through structural consistency-based preprocessing. This is a critically important factor in the whole-body virtual wearing system for which practical clothing replacement is the purpose, and this study is one of the key contribution points that dis-tinguishes it from the existing methodology.

In conclusion, CaP-VTON is a model that most accurately reflects the style and sil-houette of reference clothing, experimentally demonstrating that it provides the most reli-able synthetic quality from a user experience perspective beyond simple image quality or structural similarity indicators.

\begin{table}[H]
    \renewcommand{\arraystretch}{1.3}
    \centering
    \caption{Numerical comparison with other models. The lower the FID and LPIPS, the better the quality, and the higher the SSIM, the better the quality.}
    \begin{tabular}{l|ccc|c}
        \hline
        \textbf{Model} & \multicolumn{3}{c|}{\textbf{paired}} & \textbf{unpaired} \\
        \cline{2-4}
        & \textbf{FID $\downarrow$} & \textbf{SSIM $\uparrow$} & \textbf{LPIPS $\downarrow$} & \textbf{FID $\downarrow$} \\
        \hline
        CatVTON & 5.42 & 0.870 & 0.057 & 9.02 \\
        IDM-VTON & 5.76 & 0.850 & 0.063 & 9.84 \\
        OOTDiffusion & 9.30 & 0.819 & 0.088 & 12.41 \\
        Leffa & 4.54 & 0.899 & 0.048 & 8.52 \\
        CaP-VTON (ours) & 6.55 & 0.897 & 0.062 & 11.28 \\
        \hline
    \end{tabular}
    \label{tab:numerical_comparison}
\end{table}

\subsubsection{Comparison for Image Synthesis Image Quantity}
Table 2 compares performance between the existing virtual wearing models and the proposed CaP-VTON model, with FID, SSIM, and LPIPS indicators based on the assigned category. The proposed CaP-VTON model achieved a slightly lower quantitative performance, with FID 11.46, SSIM 0.8573, and LPIPS 0.0849, due to the addition of the image inference pipeline (Gen-erate Skin method).

When compared numerically only, our research shows moderate quality among the other synthetic models mentioned in the paper. This means that the synthesis result is not the best, and it does not produce heterogeneity to the degree to which humans perceive it. These figures are general indicators that focus on overall image quality and percep-tion-based similarity. In other words, there is a limit to evaluations of the accuracy of the transition from long-sleeve to short-sleeve, which is the purpose of our paper. Therefore, observation-based accuracy was added and evaluated.

\subsubsection{Ablation study}
Figure 10 shows the result of an ablation study analyzing the effect on the resulting quality of each major element comprising the CaP-VTON model. The same input image and reference clothing were applied in each configuration, and the short-sleeved silhou-ette and the degree of arm/shoulder skin exposure were compared in the output image. As a result, when using only Leffa, most of the existing long-sleeved clothing silhouettes re-mained present in the output image, and the exposure of the arms and shoulders was in-complete. In the case of Leffa + DressCode, the patterns and silhouettes of some reference clothing were reflected, but complete transformation was not made due to the influence of the existing long-sleeved structure. In the case of Leffa + Generate Skin, a significant part of the silhouette of the existing long-sleeved clothing remained, and the short-sleeved sil-houette was still incompletely reproduced. Finally, the CaP-VTON model of Leffa + DressCode + Generate Skin integrated all the components required to minimize the influ-ence of existing long-sleeved clothing, such that the short-sleeved silhouette and pattern of the reference clothing were naturally reflected.

\begin{figure}[H]
    \centering
    
    \begin{minipage}{0.22\linewidth}
        \centering
        \includegraphics[width=\linewidth]{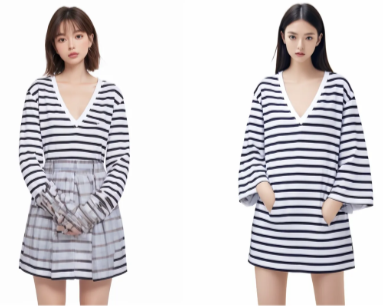}\\
        (a)
    \end{minipage}
    \hfill
    \begin{minipage}{0.22\linewidth}
        \centering
        \includegraphics[width=\linewidth]{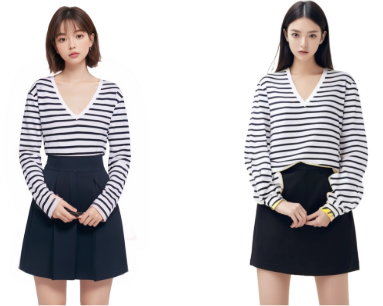}\\
        (b)
    \end{minipage}
    \hfill
    \begin{minipage}{0.22\linewidth}
        \centering
        \includegraphics[width=\linewidth]{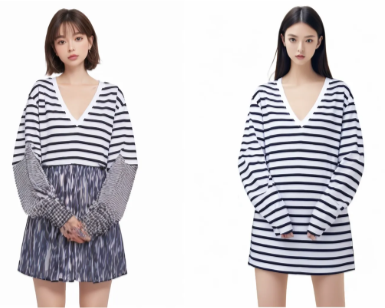}\\
        (c)
    \end{minipage}
    \hfill
    \begin{minipage}{0.22\linewidth}
        \centering
        \includegraphics[width=\linewidth]{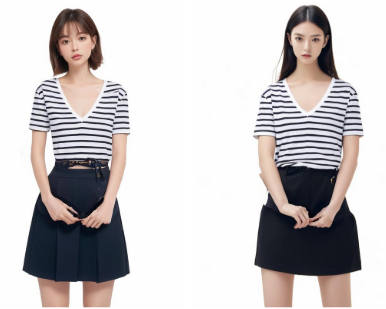}\\
        (d)
    \end{minipage}

    \caption{Ablation study consisting of (a) Leffa only, (b) Leffa + DressCode, (c) Leffa + Generate Skin, and (d) CaP-VTON of Leffa + DressCode + Generate Skin. For each configuration, the same input image and reference clothing were applied to compare the short-sleeved silhouette and the degree of arm/shoulder skin exposure in the output image.}
    \label{fig:ablation_study}
\end{figure}

\subsubsection{Evaluation based on Short-Sleeve Synthesis Probability}
Leffa + DressCode and CaP-VTON were selected to quantitatively compare the results of short-sleeved clothing synthesis by the CaP-VTON model. The reasons for this are as follows. First, as pointed out at the beginning of this paper, Leffa’s detection is limited to bottoms and by the lack of connectivity to clothing types due to modified masking, so it is not suitable for use as a comparison target when evaluating the improvement induced by the proposed model. In other words, due to structural limitations, short-sleeved clothing conversion itself is not smoothly performed by this model, making it impossible to com-pare performance under the same conditions. Second, Leffa + Generate Skin is a configu-ration in which only a skin restoration module is applied without DressCode data, and it is insufficient to evaluate clothing silhouette conversion in the virtual try-on process.

\begin{figure}[H]
    \centering
    \includegraphics[width=0.6\linewidth]{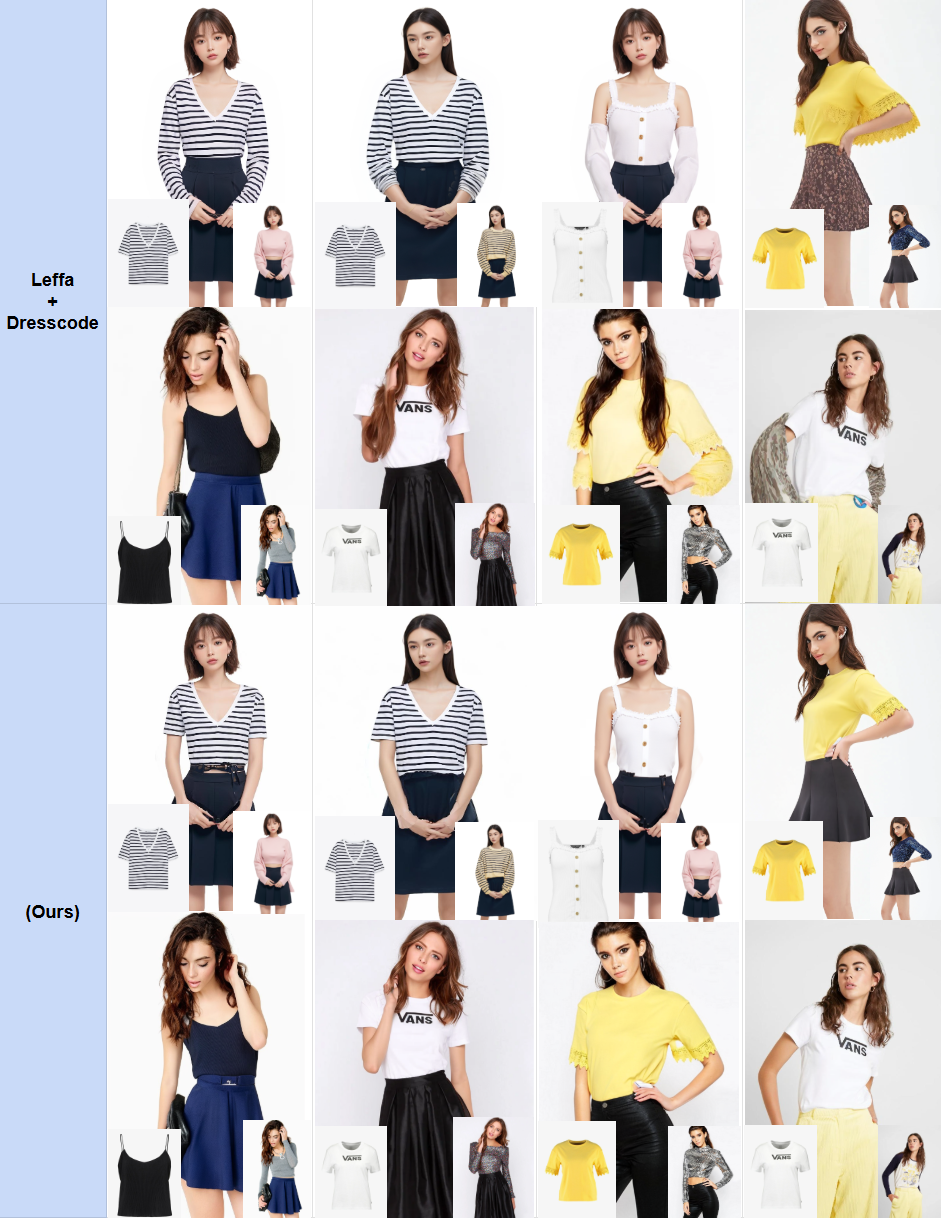}
    \caption{Comparison of short sleeve \& sleeveless clothing output}
    \label{fig:placeholder}
\end{figure}
Figure 11 is the result of comparing the output of the Leffa model and the CaP-VTON model learned with the DressCode dataset to quantitatively compare the short-sleeved clothing synthesis results of the proposed CaP-VTON model. In addition to the existing example images, real person images of the DeepFashion dataset were added to the image comparison to verify the quality of synthesis in more realistic situations.

The DeepFashion dataset is a dataset for the purpose of clothing recognition and at-tribute classification, and the purpose and characteristics of data are different from VI-TON-HD or DressCode configured for virtual try-on. Therefore, when applying Deep-Fashion data to a model that has already been sufficiently trained on VITON-HD or DressCode data, such as existing Leffa, the model tests the performance in new data and determines whether it can maintain stable performance in real person data or not as an objective evaluation index.

As can be seen from the figure, the silhouette consistency between short-sleeved and sleeveless clothing is more natural than that achieved by Leffa, and the boundary restora-tion of skin-exposed areas is smooth. In particular, when converting the long-sleeved in-put image to an output with short-sleeved clothing, it was observed that residual sleeves, parts of the body, or discontinuous textures remained on the arm in the existing Leffa, but in the proposed CaP-VTON model, the silhouette of the existing clothes was completely removed through the combination of Generate Skin pretreatment and multicategory masking, and the arm shape and skin tone were naturally restored.

In this study, CLIP-based clothing shape consistency evaluation was performed to automatically determine whether the reference short-sleeved clothing was actually syn-thesized into a normal short-sleeve image, with the intention of expanding to a long-sleeved input image. This evaluation was implemented by comparing the semantic similarity between images and texts using OpenAI's CLIP (ViT-B/32) model. Specifically, each reference vector was generated by embedding two text prompts—"A person wearing a short sleeve" and "A person wearing a long sleeve". After that, for each composite image, the visual feature vector was extracted through the image encoder of CLIP, and the Cosine Similarity between the two text vectors was calculated. Through this approach, if the composite image showed a greater similarity to the “short sleeve” prompt, it was classi-fied as normal output, and on the contrary, if it showed a greater similarity to the “long sleeve” prompt, it was classified as abnormal output. In addition, if the difference between the two similarities (score diff = correct score - wrong score) was 0.02 or more, it was classified as normal with clear consistency, and if it was less than 0.02, it was partially classified as normal.

\begin{figure}[htbp]
    \centering
    \includegraphics[width=0.8\linewidth]{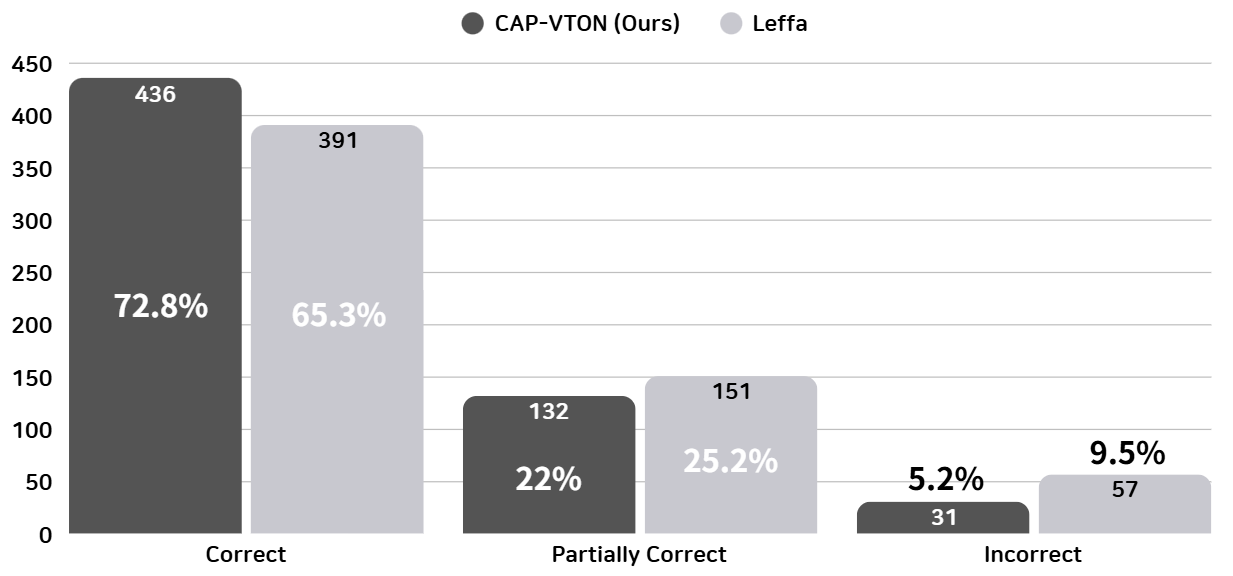}
    \caption{Comparison the percentage of correct output}
    \label{fig:placeholder}
\end{figure}

Figure 12 visualizes the classification of the output results into normal output (when a short-sleeved silhouette consistent with the reference clothing is accurately created) and abnormal output (when a long-sleeved silhouette remains, or short-sleeved clothes are synthesized as long-sleeved clothes). After that, the ratios of each evaluation grade were aggregated to calculate the normal output ratio and abnormal output ratio for each model, and these were visualized in Figure 12.

This result means that CAP-VTON recorded about a 15.4\% greater clothing silhouette consistency accuracy than Leffa. In particular, CaP-VTON showed excellent performance in its ability to accurately reflect short-sleeved reference clothing after removing long sleeves and naturally restoring the skin of the arms and shoulders, which can be said to empirically prove the effectiveness of Generate Skin-based pre-processing and the im-proved masking pipeline.

Meanwhile, Leffa showed excellent quantitative performance regarding the FID and SSIM, but the above experimental results suggest the limitation that it is difficult to fully explain the suitability of the model's clothing replacement using only quantitative image quality evaluation. In practical applications, the accuracy and visual naturalness of the clothing shape transformation desired by the user are more important factors, and thus, the consistency-based evaluation method proposed in this study is a very important measure in a practical sense.

In conclusion, CaP-VTON showed a marked improvement over Leffa regarding the structural consistency, meaning that the pipeline was designed to minimize reliance on existing clothing types and accurately reflect the style of reference clothing. These results support the feasibility of using CaP-VTON in high-dimensional applications such as full-body virtual wear and style transformation.

\section{Discussion}
\subsection{Improved entire pipeline}
The improved masking pipeline and Generate Skin-based skin inpainting pipelines do not just complement each function, but also create complementary synergies at the en-tire pipeline level. In particular, the integration of these two modules effectively alleviated the structural limitations of the existing Leffa model, and produced stable and natural re-sults, especially when long/short-sleeve conversion or partial skin exposure correction was required.

Since the existing Leffa model specialized in top-centered synthesis, it showed a lim-itation in not being able to properly handle complex clothing structures such as bottoms or dress categories. These limitations result in two key problems: the incompleteness of detection of clothing areas and the persistence of existing clothing silhouettes. Therefore, this study introduced a double complementary pipeline structure that considers both the improvement of masking accuracy and the effect of removing existing clothing.

The masking pipeline utilizes multi-category information from the DressCode da-taset to perform accurate clothing boundary detection, and based on the results, generates an appropriate inpainting mask to remove existing clothing areas. Subsequently, the gen-erate skin pipeline performs a function of blocking existing clothing characteristics in-cluded in the input image from interfering with the final output by realistically restoring the skin and short-sleeved silhouette to the removed area. This dual processing structure provides an ideal condition for accurately and consistently reflecting the shape of the ref-erence clothing, regardless of the shape of the existing clothing.

For example, in a scenario wherein a person wearing a long-sleeve shirt wears a sleeveless dress, the masking pipeline accurately detects the long-sleeve area, from the shoulder to the wrist, and the Skin Inpainting Pipeline naturally restores the skin therein, thereby accurately reproducing the exposure design of the reference dress. This is not just a partial modification, but a structural improvement of the entire process, leading to pre-masking–inpainting–diffusion for whole-body synthesis.

\subsection{Supporting for replacement across different clothing types}
This integrated pipeline is designed based on the multi-category structure of the DressCode dataset, enabling free clothing replacement between two categories, namely, top/bottom, top/dress, and bottom/dress. This function was technically impossible in the top-centered synthesis structure of the existing Leffa model, and it is difficult to implement without the structural improvement proposed in this study. In addition, for structurally complex clothing in which the top and bottom are integrated, such as dresses, bottom masking and upper body skin restoration are applied in parallel, enabling natural and flawless synthesis results. This allows users to produce complex styles that include not only style changes within a single category, but also shape transitions between categories, which greatly expands the scope of use of virtual wearing systems.

\subsection{Clothing-agnostic properties}
The existing virtual try-on models have limitations, such as not being able to com-pletely remove the silhouette of the clothes in the input image, or being synthesized into the bottom area. For example, when short-sleeved clothes are synthesized on a person wearing a long-sleeved shirt, the arm outline of the original image remains very small, resulting in unnatural overlapping in the resulting image.

In this study, to address these limitations, the existing clothing silhouette was re-moved using a skin inpainting pipeline based on the DressCode masking pipeline and Generate Skin method, and the natural arm and skin areas suitable for the body structure of the original person were restored. Through this, regardless of what type of clothing the input image features, a clothing-agnostic model that can synthesize new reference cloth-ing while the clothing silhouette of the original image is completely removed has been implemented. This approach provides more realistic and natural results even in the future synthesis of skin-exposing clothing such as short-sleeves and sleeveless clothes. Therefore, the cloning-agnostic design of this study serves as an important basis for enabling stable clothing conversion, even in various exposure forms or pose deformation situations, in the future

\section{Conclusion}
This study has proposed a new pipeline structure, CaP-VTON (\textbf{C}lothing \textbf{A}gnostic \textbf{P}re-Inpainting \textbf{V}irtual \textbf{T}ry-\textbf{On}), to overcome the limitations of the existing Leffa-based virtual try-on system. While Leffa showed excellent performance in detailed texture repre-sentation through attention-based flow field learning, it had structural limitations such as bottom detection inaccuracy and the persistence of existing clothing silhouettes. To solve these two key problems, this study designed a dual complementary pipeline that inte-grates (1) a multi-category masking pipeline based on DressCode and (2) a Generate Skin preprocessing mechanism based on Stable Diffusion.

The masking pipeline has been improved to accurately detect various clothing cate-gories such as top, bottom, and dress, and the skin inpainting pipeline is configured to remove existing clothing areas and to naturally restore skin and short-sleeved silhouettes to accurately reflect the style of the reference clothing. In particular, the Generate Skin method contributes to stable transformation from long-sleeved to short-sleeved or sleeve-less clothing by naturally restoring skin areas such as arms, shoulders, and armpits, ac-cording to the body's posture, after removing clothing. The experiment showed that CaP-VTON had the same quantitative index as the existing Leffa, and at the same time, achieved an accuracy of 92.5\% in the evaluation of consistency based on the accuracy of short-sleeved silhouette synthesis, achieving a 15.4\% higher performance than Leffa's 77.1\%. In addition, it was confirmed that CaP-VTON consistently reproduces the sleeve length, pattern, and style of reference clothing, unlike other existing models.

This study makes the following contributions to the virtual inspection model. First, the bottom and dress masking functions were successfully implemented to support whole-body synthesis. Second, an input-independent synthesis structure was established by removing the interference of existing clothing. Finally, the model-agnostic modular structure secured a versatility that made it applicable to various diffusion-based models. This model thus represents a foundational technology in building a more precise and practical virtual start-up system in the actual e-commerce environment in the future, and has the potential to expand into a variety of fields such as the simulation of customized clothes, avatar generation, and clothing design previews. Future research is expected to further expand the practicality and applicability of the CaP-VTON pipeline by further elaborating the processing of complex poses, backgrounds, and interactions with non-clothing elements such as accessories and hair; improving inference speed in re-al-time environments; and integrating all this with a dynamic clothing recommendation system based on user feedback.



\clearpage

\section{References}
\begin{thebibliography}{99}

\bibitem{Zhou2025} Zhou, Z.; Liu, S.; Han, X.; Liu, H.; Ng, K.W.; Xie, T.; Cong, Y.; Li, H.; Xu, M.; Pérez-Rúa, J.M.; Patel, A.; Xiang, T.; Shi, M.; He, S. Learning Flow Fields in Attention for Controllable Person Image Generation. In Proceedings of the IEEE/CVF Conference on Computer Vision and Pattern Recognition (CVPR), Nashville, TN, USA, 2025; pp. 2491-2501. \url{https://doi.org/10.1109/CVPR52734.2025.00238}

\bibitem{Heusel2017} Heusel, M.; Ramsauer, H.; Unterthiner, T.; Nessler, B.; Hochreiter, S. GANs Trained by a Two Time-Scale Update Rule Converge to a Local Nash Equilibrium. Advances in Neural Information Processing Systems (NeurIPS), 2017, vol. 30, pp. 6626-6637. arXiv:1706.08500. \url{https://doi.org/10.18034/ajase.v8i1.9}

\bibitem{Wang2004} Wang, Z.; Bovik, A.C.; Sheikh, H.R.; Simoncelli, E.P. Image Quality Assessment: From Error Visibility to Structural Similarity. IEEE Transactions on Image Processing, 2004, 13(4), pp. 600-612. \url{https://doi.org/10.1109/TIP.2003.819861}

\bibitem{Zhang2018} Zhang, R.; Isola, P.; Efros, A.A.; Shechtman, E.; Wang, O. The Unreasonable Effectiveness of Deep Features as a Perceptual Metric. In Proceedings of the IEEE/CVF Conference on Computer Vision and Pattern Recognition (CVPR), Salt Lake City, UT, USA, 2018; pp. 586-595. \url{https://doi.org/10.1109/CVPR.2018.00068}

\bibitem{WangECCV2018} Wang, B.; Zheng, H.; Liang, X.; Xhen, Y.; Lin, L. Toward Characteristic-Preserving Image-Based Virtual Try-On Network (CP-VTON). In Computer Vision—ECCV 2018, Part XIII; Springer: Cham, Switzerland, 2018; pp. 607-623. \url{https://doi.org/10.1007/978-3-030-01261-8_36}

\bibitem{Choi2021} Choi, S.; Park, S.; Lee, M.; Choo, J. VITON-HD: High-Resolution Virtual Try-On via Misalignment-Aware Normalization. In Proceedings of the IEEE/CVF Conference on Computer Vision and Pattern Recognition (CVPR), Nashville, TN, USA, 2021; pp. 14131-14140. \url{https://doi.org/10.1109/CVPR46437.2021.01391}

\bibitem{Yang2020} Yang, H.; Min, S.; Chen, X.; Wang, Y.; Lin, L. Towards Photo-Realistic Virtual Try-On by Adaptively Generating-Preserving Image Content (ACGPN). In Proceedings of the IEEE/CVF Conference on Computer Vision and Pattern Recognition (CVPR), Seattle, WA, USA, 2020; pp. 11854-11863. \url{https://doi.org/10.1109/cvpr42600.2020.00787}

\bibitem{LeeECCV2022} Lee, J.; Kim, S.; Kim, D.; Sohn, K. HR-VITON: High-Resolution Virtual Try-On via Joint Layout and Texture Learning. In Computer Vision—ECCV 2022; Springer: Cham, Switzerland, 2022; pp. 280-296. \url{https://doi.org/10.1007/978-3-031-19790-1_13}

\bibitem{Goodfellow2014} Goodfellow, I.; Pouget-Abadie, J.; Mirza, M.; Xu, B.; Warde-Farley, D.; Ozair, S.; Courville, A.; Bengio, Y. Generative Adversarial Nets. Advances in Neural Information Processing Systems (NeurIPS), 2014, vol. 27, pp. 2672-2680. \url{https://doi.org/10.1145/3422622}

\bibitem{Kim2024} Kim, J.; Gu, G.; Park, M.; Park, S.; Choo, J. StableVITON: Learning Semantic Correspondence with Latent Diffusion Model for Virtual Try-On. In Proceedings of the IEEE/CVF Conference on Computer Vision and Pattern Recognition (CVPR), Seattle, WA, USA, 2024; pp. 8176-8185. \url{https://doi.org/10.1109/cvpr52733.2024.00781}

\bibitem{Li2024} Li, Z.; et al. IDM-VTON: Improving Diffusion Models for Authentic Virtual Try-On in the Wild. arXiv 2024, arXiv:2403.05139. \url{https://doi.org/10.48550/arXiv.2403.05139}

\bibitem{Morelli2023} Morelli, D.; et al. LaDI-VTON: Latent Diffusion Textual-Inversion Enhanced Virtual Try-On. In Proceedings of the ACM International Conference on Multimedia (ACM MM), 2023. \url{https://doi.org/10.1145/3581783.3612137}

\bibitem{Xu2024} Xu, Y.; Gu, T.; Chen, W.; Chen, C. OOTDiffusion: Outfitting Fusion Based Latent Diffusion for Controllable Virtual Try-On. arXiv 2024, arXiv:2403.01779. \url{https://doi.org/10.48550/arXiv.2403.01779}

\bibitem{Vaswani2017} Vaswani, A.; et al. Attention Is All You Need. Advances in Neural Information Processing Systems (NeurIPS), 2017. \url{https://doi.org/10.48550/arXiv.1706.03762}

\bibitem{Chong2025} Chong, Z.; et al. CatVTON: Concatenation Is All You Need for Virtual Try-On with Diffusion Models. In Proceedings of the International Conference on Learning Representations (ICLR), 2025. \url{https://doi.org/10.48550/arXiv.2407.15886}

\bibitem{Zhu2023} Zhu, L.; Yang D.; Zhu, T.; Reda , F.; Chan, W.; Saharia, C. ; Norouzi, M.; Kemelmacher-Shlizerman I. Tryondiffusion: A tale of two unets. In CVPR. 2023; pp.4606-4615. https://arxiv.org/abs/2306.08276

\bibitem{Zhang2023-1} Zhang, K.;Sun, M.; Sun, J.; Zhao, B.; Zhang, K.; Sun, Z.; Tan, T. HumanDiffusion: A coarse-to-fine alignment diffusion framework for controllable text-driven person image generation, 2023, https://arxiv.org/abs/2211.06235

\bibitem{Li2021} Li, P.; Song, G.; Zhang, Y.; Tong, Z.; Wei, X.; Liu, Y.; Yang, X. Self-Correction for Human Parsing. IEEE Trans. Pattern Anal. Mach. Intell., 2021, 43, 4213–4225. \url{https://doi.org/10.1109/TPAMI.2020.3048039}

\bibitem{Cao2017} Cao, Z.; et al. OpenPose: Realtime Multi-Person 2D Pose Estimation Using Part Affinity Fields. In Proceedings of the IEEE/CVF Conference on Computer Vision and Pattern Recognition (CVPR), 2017; pp. 7291–7299. \url{https://doi.org/10.1109/CVPR.2017.143}

\bibitem{Zhang2023-2} Zhang, L.; Agrawala, M. Adding Conditional Control to Text-to-Image Diffusion Models. In Proceedings of the IEEE/CVF International Conference on Computer Vision (ICCV), 2023; pp. 3813–3824. \url{https://doi.org/10.1109/ICCV51070.2023.00355}

\bibitem{Merjic2023} Merjic. majicMIX Realistic v7: Stable Diffusion Checkpoint Merge. HuggingFace, 2023. \url{https://huggingface.co/imagepipeline/MajicMIX-realistic}

\end{thebibliography}
\end{document}